%% file: main.tex
\documentclass[10pt,twocolumn,letterpaper]{article}

\usepackage{titlesec}

\usepackage{cvpr}              
\usepackage{textcomp}
\usepackage{stfloats}
\usepackage{url}
\usepackage{verbatim}
\usepackage{graphicx}
\usepackage{cite}
\usepackage{amssymb}
\usepackage{amsmath,amsfonts}
\usepackage{pifont}
\usepackage{multirow}
\usepackage{booktabs}
\usepackage{float}
\usepackage{bm}
\usepackage{gensymb}
\usepackage{comment}
\usepackage{tabularx}
\usepackage{mathtools}

\usepackage{pifont}

\newcommand{\tick}{\ding{51}} 

\usepackage[table]{xcolor} 
\definecolor{lightgray}{gray}{0.8} 

\newcolumntype{Y}{>{\centering\arraybackslash}X}

\definecolor{cvprblue}{rgb}{0.21,0.49,0.74}
\usepackage[pagebackref,breaklinks,colorlinks,citecolor=cvprblue]{hyperref}

\graphicspath{{./figures/}}

\usepackage{balance}

\usepackage{xspace}
\usepackage{xstring}

\newcommand{\codestyle}[1]{\texttt{#1}\xspace}
\def\ln{\codestyle{LN}}
\def\tsmb{\codestyle{TSMB}}
\def\pgmb{\codestyle{PGMB}}
\def\pgml{\codestyle{PGM}}
\def\tdown{\codestyle{TDown}}
\def\mysplit{\codestyle{Split}}
\def\myconcat{\codestyle{Concat}}
\def\linear{\codestyle{Linear}}
\def\tconv{\codestyle{TemporalConv}}
\def\gconv{\codestyle{SpatialConv}}
\def\bn{\codestyle{BN}}
\def\ffn{\codestyle{FFN}}

\def\convoned{\codestyle{Conv1D}}

\def\gelu{\codestyle{GELU}}
\def\ssm{\codestyle{SSM}}
\def\mycssm{\codestyle{C2DSSM}}
\def\pool{\codestyle{Pool}}

\def\yhat{\mathbf{\hat{y}}}

\def\modality{\mathbb{M}}
\def\modalityj{\modality_j}
\def\modalityjb{\modality_{jb}}
\def\modalityjbm{\modality_{jbm}}

\titlespacing*{\subsubsection}{\parindent}{0pt}{0pt}

\titleformat{\subsubsection}[runin]
  {\normalfont\normalsize\bfseries}
  {}
  {0pt}
  {}
  [] %

\def\paperID{7217} 
\def\confName{CVPR}
\def\confYear{2025}

\title{
\begin{tabularx}{\linewidth}{l X r}
\raisebox{-2em}[0pt][0pt]{\hspace{2em}\includegraphics[height=3em]{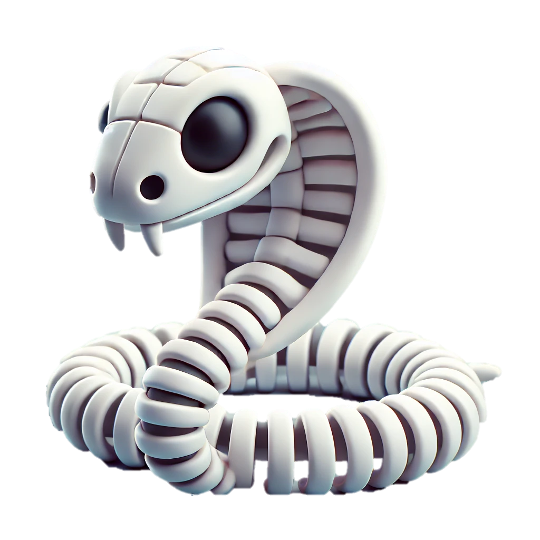}} &
\centering SkelMamba: A State Space Model for Efficient Skeleton Action Recognition of Neurological Disorders &
\raisebox{-2em}[0pt][0pt]{\includegraphics[height=3em]{figures/mamba_skel_icon.png}}
\\
\end{tabularx}
}

\author{Niki Martinel\\
University of Udine\\
{\tt\small niki.martinel@uniud.it}
\and
Mariano Serrao\\
Sapienza Università di Roma\\
{\tt\small mariano.serrao@uniroma1.it}
\and
Christian Micheloni\\
University of Udine\\
{\tt\small christian.micheloni@uniud.it}
}

\begin{document}
\maketitle
\input{sec/0_abstract}    
\input{sec/1_intro}

\input{sec/2_related_work}

\input{sec/3_method}

\input{sec/4_exp}
\input{sec/5_ablation}
\input{sec/6_conclusion}

{
    \small
    \bibliographystyle{ieeenat_fullname}
    \bibliography{main}
}

\end{document}

%% file: sec/0_abstract.tex
\begin{abstract}
We introduce a novel state-space model (SSM)-based framework for skeleton-based human action recognition, with an anatomically-guided architecture that improves state-of-the-art performance in both clinical diagnostics and general action recognition tasks.
Our approach decomposes skeletal motion analysis into spatial, temporal, and spatio-temporal streams, using channel partitioning to capture distinct movement characteristics efficiently.
By implementing a structured, multi-directional scanning strategy within SSMs, our model captures local joint interactions and global motion patterns across multiple anatomical body parts. This anatomically-aware decomposition enhances the ability to identify subtle motion patterns critical in medical diagnosis, such as gait anomalies associated with neurological conditions. 
On public action recognition benchmarks,~\ie, NTU RGB+D, NTU RGB+D 120, and NW-UCLA, our model outperforms current state-of-the-art methods, achieving accuracy improvements up to 3.2\% with lower computational complexity than previous leading transformer-based models.
We also introduce a novel medical dataset for motion-based patient neurological disorder analysis to validate our method’s potential in automated disease diagnosis.
\end{abstract}

%% file: sec/1_intro.tex
\section{Introduction}
\label{sec:intro}
Human action recognition is the task of classifying actions based on human movements. 
The problem is often tackled by leveraging the rich contextual features in RGB videos --at the cost of exposing information about people's identities.
Skeleton-based action recognition has emerged as a privacy-preserving alternative for sensitive applications, from patient monitoring and physical therapy to assisted living environments.

Skeletal 3D joint representations are compact and robust to environmental conditions (\eg, background clutter and light variations) yet their sparse nature makes this task inherently challenging.
In the medical domain, precisely capturing the dynamic spatio-temporal relationships between joints is of paramount importance for precise analysis of subtle movements indicative of various diseases.
For example, analyzing a patient's gait can provide insights into neurological disorders, musculoskeletal abnormalities, and other health conditions.

The skeleton joints and their connections (\ie, bones) correspond to the vertices and edges of a graph structure. 
Our community has recently proposed skeleton action recognition methods that model the spatial and temporal dependencies among skeletal joints via Graph Convolution Networks (GCN) or transformer-like architectures.
GCN-based methods introduced adaptive graph structures (\eg,~\cite{shi2019dgnn, chi2022infogcn, shi2019agcn}), specialized joint encodings (\eg,~\cite{lin2023actionlet, cai2021jologcn}), and explored multiple modalities (\eg,~\cite{liu2024mmcolearning}) for learning robust representations (\eg,~\cite{hou2022graphmae,zhou2023,lin2023actionlet}).
Transformer-based methods tackle the long-range dependencies in skeletal data that GCN-based methods often struggle with.
Existing methods model the skeletal spatio-temporal relationship between physically neighboring and distant joints/frames using the self-attention mechanism (\eg,~\cite{zhou2022hyperformer, do2024skateformer}) in one-shot settings (\eg,~\cite{zhu2023adaptive}) or for joint training across different action tasks and datasets (\eg,~\cite{duan2023skeletr}).

Transformer-like architectures are computationally demanding and GCN-based methods struggle to model the relation of physically distant joints --captured through the direct propagation of information between physically connected joints.
This motivates the introduction of our novel state-space model (SSM)-based architecture, which models all joint relationships with an efficient spatial-temporal scanning strategy --designed to analyze skeletal data for patient disease recognition.

Our novel approach introduces a structured decomposition of skeletal motion data operating across three complementary dimensions. Given an input sequence, we first partition its channel representations into specialized groups for spatial, temporal, and spatio-temporal analysis.
The spatial and temporal streams capture local patterns and short-range frame-wise transitions, while the spatio-temporal stream introduces State Space Models (SSMs) for complex motion modeling.

Neurological disorders impact distinct body parts during locomotion, resulting in disease-specific motion patterns. 
Within the spatio-temporal stream, we further partition the input according to anatomically meaningful body parts (\eg, legs, torso, arms) and their key interactions (\eg, arms-legs coordination) that are analyzed by separate SSMs.
Each SSM features our novel four-way scanning strategy that splits each anatomical group into four channel subgroups.
A specific scanning direction is applied to a specific subgroup.
This enables efficient parallel processing --while reducing the computational demands of our model-- and allows us to collectively analyze the motion patterns across both space and time: from temporal to spatial domain, spatial to temporal domain, and their respective inverse directions.
This multi-directional scanning enables comprehensive capture of both local joint relationships and global motion patterns while being computationally efficient.

This new anatomically-aware architecture proves particularly effective for automated medical diagnosis, where subtle motion abnormalities often manifest through complex interactions between different body parts over time.
Nevertheless, our model remains generic and demonstrates significant improvements over state-of-the-art results on existing challenging action recognition datasets, showcasing its versatility and robustness.

Our contribution is threefold:
\begin{itemize}
\item We propose a novel multi-stream architecture leveraging SSMs that effectively decomposes motion analysis into spatial, temporal, and spatio-temporal streams through channel partitioning, enabling efficient parallel processing of distinct motion characteristics.
\item We introduce an anatomically-aware partitioning scheme that guides the SSM analysis based on meaningful body parts and their interactions, capturing both local joint dynamics and complex cross-body motion patterns crucial for medical diagnosis.
\item We develop a channel-split scanning mechanism where input features are partitioned into four subgroups, each processed by direction-specific SSMs. This approach enables comprehensive multi-directional motion analysis while maintaining computational efficiency through reduced channel dimensionality.
\end{itemize}

Through extensive experiments on both medical diagnosis tasks (we introduce a dataset for the analysis of patient walking styles to assist in automated neurological disease diagnosis) and standard action recognition benchmarks, we demonstrate that our method achieves state-of-the-art performance while being very computationally efficient.

%% file: sec/2_related_work.tex
\section{Related Work}
\subsubsection{Graph Convolutional Networks (GCNs)}~have been initially explored for skeleton-based action recognition in~\cite{yan2018stgcn}.
The seminal work introduced the Spatial-Temporal Graph Convolutional Network (ST-GCN) framework for modeling human joints with spatiotemporal graph structures.
Recent architectural innovations have evolved along multiple dimensions, from joint-bone fusion networks \cite{peng2020gnnsearch} and multi-scale feature extraction \cite{shi2020msagcn} to adaptive graph topology learning \cite{shi2019agcn}.
Context-aware architectures \cite{zhang2020contextawaregcn, chen2021ctrgcn} and dual-stream temporal models \cite{shi2020dstanet} have enhanced feature extraction capabilities, while adaptive graph structures \cite{myung2024degcn, ye2020dyngcn} and efficient convolutions \cite{cheng2020shiftgcn, duan2022pyskl} have optimized computational overhead.
These developments, complemented by multi-modal integration approaches \cite{liu2024mmcolearning}, have significantly advanced the field's ability to capture rich motion characteristics while maintaining robustness across varying conditions.

GCN-based approaches rely on local graph operations and predefined adjacency matrices, limiting their ability to capture long-range dependencies and dynamic motion patterns. 
Our approach diverges by introducing state-space models (SSMs) for dynamic latent space partitioning, efficiently capturing local and global interactions.

\subsubsection{Transformers}~are a powerful alternative architecture for skeleton-based action recognition, primarily due to their capacity to model complex joints and temporal motion patterns through the self-attention mechanism \cite{vaswani2017attention}.
Spatial-temporal attention frameworks \cite{plizzari2021sttr, duan2023skeletr} enable joint modeling of structural and dynamic information, while global-local attention mechanisms \cite{kim2022globallocaltransformer} selectively focus on key motion patterns across different temporal scales.
Frequency-aware architectures \cite{wu2024freqaware} improve data efficiency through spectral augmentation, while specialized designs \cite{do2024skateformer} optimize performance for specific motion types.
Multi-modal approaches \cite{wang20233mformer} incorporate complementary sensor data to enhance robustness across varying conditions.
Self-supervised pretraining strategies \cite{zhou2022hyperformer} leverage unlabeled data for improved representation learning, while efficient architectural designs \cite{pang2022igformer, oraki2024lortsar} maintain high accuracy with reduced computational demands.

Despite these advances, transformer-based methods face inherent limitations due to their quadratic computational complexity in modeling pairwise attention, motivating our exploration of more efficient architectures for real-time applications.

\subsubsection{State-space models (SSMs)}~offer an efficient approach to sequence modeling with long-range dependencies. Initial work on linear state-space layers~\cite{gu2021combining} laid the foundation for S4~\cite{gu2022s4}, while subsequent variants~\cite{gupta2022diagonalssm,smith2023simplifiedssm,hasani2022liquidssm} demonstrated comparable performance with simplified architectures.
Mamba~\cite{guo2023ssmmamba} addressed content-based reasoning limitations through input-dependent parameters~\cite{fu2022hungry}.
For vision tasks, existing SSM adaptations~\cite{nguyen2022s4nd,ma2024umamba,archit2024vimunet,patro2024simba} primarily follow fixed unidirectional scanning patterns.
Video-based methods~\cite{yang2024vivim,zhang2024motionmamba,chen2024videomamba} process temporal information sequentially after spatial encoding, while 3D approaches~\cite{zhang2024voxelmamba,liang2024pointmamba,shaker2024groupmamba} rely on predetermined serialization strategies. 
Recent works have further expanded SSMs to image restoration~\cite{deng2024cumamba} and speech processing~\cite{liu2024spmamba}, with surveys~\cite{xu2024mambasurvey} providing comprehensive overviews.

While SSMs offer flexibility, their adaptation to skeleton-based action recognition remains challenging.
The recent work in~\cite{chaudhuri2024simba} addressed the problem with a temporal-driven one-direction scanning strategy applied on the top of GCN latent space embeddings.
In contrast, we enable simultaneous four-way scanning  (spatial-temporal, temporal-spatial, in both forward/backward directions) through channel grouping.
Experimental comparisons show that this efficient multi-directional processing captures richer motion dependencies yielding improved skeleton-based analysis.

\begin{figure*}[!t]
\centering
\includegraphics[width=1\linewidth]{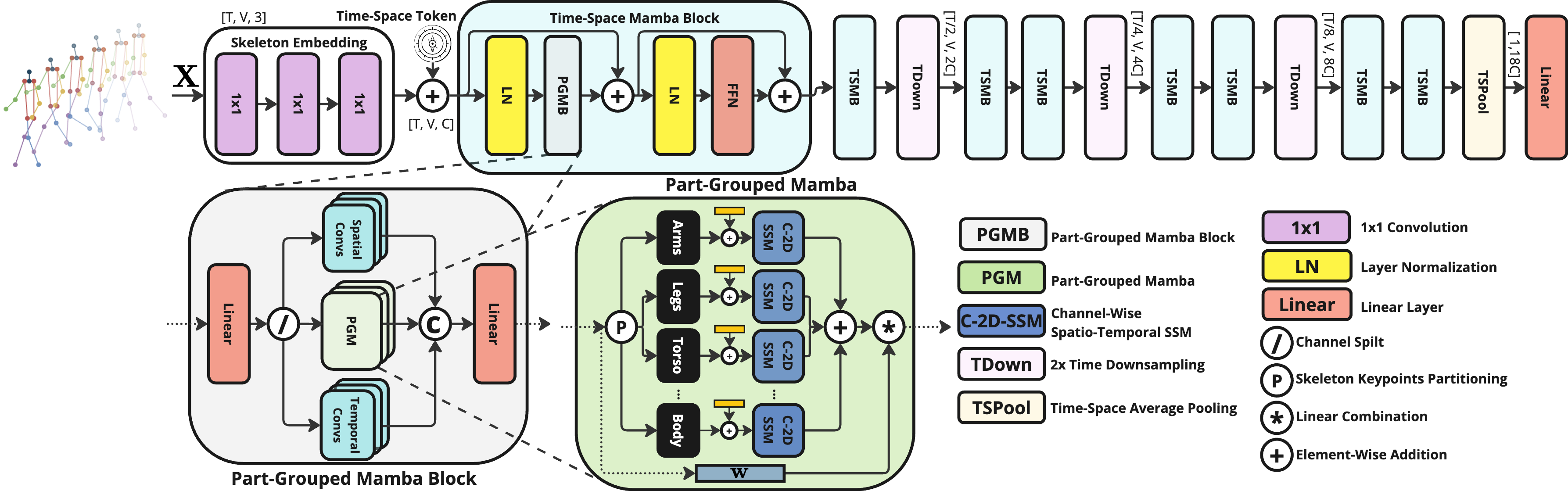}
\caption{The overall framework of our proposed SkelMamba architecture.}
\label{fig:framework}
\end{figure*}

\subsubsection{Automated diagnosis of neurological disorders}~has made significant progress through multi-modal approaches and advanced architectures.
Recent methods have demonstrated success by integrating skeleton data with foot pressure information for Parkinson's Disease (PD) assessment \cite{naseem2024gaitfootpressureskeleton}, while graph-based networks enhanced with causality mechanisms have improved Freezing of Gait detection \cite{guo2024causalitygcn}.
Clinical findings have shown that fine-tuned motion encoders effectively capture pathological gait patterns \cite{adeli2024benchmarkingskeletonparkinson}, complemented by spatiotemporal architectures for PD recognition \cite{zhang2023wmstgcn}.
Vision-based ensemble discriminate between PD and knee osteoarthritis gaits \cite{kour2024koanmpddataset}, building upon established clinical characterizations of disorder-specific gait patterns \cite{10.1093/brain/122.7.1349, Serrao2016}.
More recently, transformer architectures have demonstrated promising results in early PD detection \cite{ma2024twintower}.
These advances underscore the growing potential of computer vision for objective clinical gait assessment \cite{guo2022parkinsongaitsurvey}.

These existing approaches often hinge on complex preprocessing/modeling architecture or raise privacy concerns through video analysis.
Our model operates directly on skeletal data and captures distinct spatiotemporal dynamics in walking patterns without requiring extensive computational resources or compromising patient privacy, advancing the state-of-the-art in skeleton-based action recognition.

%% file: sec/3_method.tex
\section{SkelMamba}
\label{sec:method}

The SkelMamba architecture is illustrated in~\figurename~\ref{fig:framework}.
A skeleton sequence is denoted as $\mathbf{X} \in \mathbb{R}^{T \times V \times C}$, where $T$ is the sequence length, $V$ is the number of joints per frame, and $C$ represents the joint coordinates.

Three linear layers project the low-dimensional skeleton data onto a higher-dimensional embedding space.
A learnable time-space token is added to the embedding, then input to $L$ Time-Space Mamba Blocks (\tsmb), each containing a Part-Group Mamba Block (\pgmb) modeling the skeletal-temporal relations and a feed-forward network (\ffn) for feature refinement.
To retain temporal dynamics information while reducing the computational complexity, after two \tsmb blocks, a \tdown layer consisting of a \convoned with a stride of 2 followed by \bn is applied.
The features obtained after the $L$ {\tsmb}s undergo skeletal-temporal average pooling followed by a \linear layer producing $\yhat \in \mathbb{R}^{\mathcal{C}}$, with $\mathcal{C}$ denoting the number of classes.

\subsection{Time-Space Mamba Block (\tsmb)}
To design our novel \tsmb, we follow a similar structure to traditional transformer blocks~\cite{vaswani2017attention}.
The first part of the block models the spatial-temporal dynamics via part-grouped interactions
\begin{equation}
\mathbf{X} = \mathbf{X} + \pgmb(\ln(\mathbf{X}))
\end{equation}
with the \pgmb block computing
\begin{align}
[\mathbf{X}_{s}, \mathbf{X}_{m}, \mathbf{X}_{t}] &= \mysplit(\linear(\mathbf{X})) \\
\mathbf{X}_{s} &= \gconv(\mathbf{X}_{s}) \\
\mathbf{X}_{m} &= \pgml(\mathbf{X}_{m}) \\
\mathbf{X}_{t} &= \tconv(\mathbf{X}_{t}) \\
\mathbf{X} &= \linear(\myconcat(\mathbf{X}_{s}, \mathbf{X}_{m}, \mathbf{X}_{t}))
\end{align}
where \ln is the Layer Normalization operator, and \mysplit and \myconcat are the channel splitting and concatenation functions.
For input $\mathbf{X}$, we split it into $\mathbf{X}_{s}$, $\mathbf{X}_{m}$, and $\mathbf{X}_{t}$ with $C/4$, $C/2$, and $C/4$ channels, respectively.

Following a similar spirit to multi-head attention mechanisms, we have $H/4$ parallel \pgml, \gconv, and \tconv operators, with $H$ denoting the number of heads.
The novel \pgml layer models spatial-temporal relationships of different body parts in $\mathbf{X}_{m}$.
Each \gconv is a one-layer GCN with a learnable $(H/4, V, V)$ matrix --instead of a predefined adjacency matrix-- that captures diverse joint spatial connectivity patterns in $\mathbf{X}_{s}$.
To model temporal dynamics of patterns in $\mathbf{X}_{t}$, every \tconv performs a $H/4$-grouped \convoned filtering operation, with kernel size $k_t$.

The second part of the block refines the captured skeletal spatial-temporal dynamics by computing 
\begin{equation}
\mathbf{X} = \mathbf{X} + \underbrace{\linear(\gelu(\linear}_{\ffn}(\ln(\mathbf{X}))))
\end{equation}
which corresponds to the output of a \tsmb.

\subsection{Part-Grouped Mamba (\pgml)}
This layer is designed following three key insights:
(i) spatial (\gconv) and temporal (\tconv) operators model short-range temporal dynamics of the whole body;
(ii) different diseases affect specific body parts which exhibit distinct yet interrelated long-range temporal dynamics.
Following such intuitions, our novel \pgml introduces channel-wise driven scanning and part-based decomposition strategies in State Space Models~\cite{gu2022s4} to efficiently capture long-range local and global motion patterns.

\subsubsection{State Space Models (SSMs)}~are designed to map a 1D input sequence $x(t) \in \mathbb{R}$ to an output $y(t) \in \mathbb{R}$ using a hidden state $\mathbf{h}(t) \in \mathbb{R}^N$. Formally, this mapping is governed by the following ordinary differential equations (ODEs):
\begin{equation}
\label{eq:continuous_ssm}
\begin{aligned}
\mathbf{h}'(t) &= \mathbf{A}\mathbf{h}(t) + \mathbf{B}x(t) \\
y(t) &= \mathbf{C}\mathbf{h}(t)
\end{aligned}
\end{equation}
where \(\mathbf{A} \in \mathbb{R}^{N \times N}\) is the system’s evolution matrix, and \(\mathbf{B} \in \mathbb{R}^{N \times 1}\), \(\mathbf{C} \in \mathbb{R}^{1 \times N}\) are projection matrices.
Modern SSMs~\cite{gu2022s4} discretize~(\ref{eq:continuous_ssm}) using the zero-order hold (ZOH) method
\begin{equation}
\begin{aligned}
\mathbf{\overline{A}} &= \exp(\Delta \mathbf{A}) \\
\mathbf{\overline{B}} &= (\Delta \mathbf{A})^{-1} (\exp(\Delta \mathbf{A}) - \mathbf{I}) \cdot \Delta \mathbf{B}
\end{aligned}
\end{equation}
 with timescale parameter $\Delta$ --which can be seen as the resolution of the continuous input $x(t)$-- leading to the discrete state-space equations
\begin{equation}
\begin{aligned}
\mathbf{h}_t &= \mathbf{\overline{A}}\mathbf{h}_{t-1} + \mathbf{\overline{B}}x_t \\
y_t &= \mathbf{C}\mathbf{h}_t
\end{aligned}
\end{equation}
that can be efficiently computed by the convolution
\begin{equation}
    \begin{aligned}
\overline{\mathbf{K}} &= (\mathbf{C}\mathbf{\overline{B}}, \mathbf{C}\mathbf{\overline{A}}\mathbf{\overline{B}}, \cdots, \mathbf{C}\mathbf{\overline{A}}^{L-1}\mathbf{\overline{B}}) \\
\mathbf{y} &= \mathbf{x} * \mathbf{\overline{K}}
\end{aligned}
\end{equation}
where $L$ denotes the length of the input sequence $\mathbf{x}$ and $\mathbf{\overline{K}}$ is the SSM convolutional kernel.

Unlike traditional linear time-invariant SSMs, Mamba~\cite{guo2023ssmmamba} introduces a Selective Scan Mechanism (S6) such that parameters $\mathbf{B}$, $\mathbf{C}$, and $\Delta$ are directly derived from the input data, hence allowing input-dependent interactions along the sequence.

\subsubsection{Channel-Wise Spatio-Temporal SSM (C-2D-SSM):}
\begin{figure}[t]
\centering
\includegraphics[width=\linewidth]{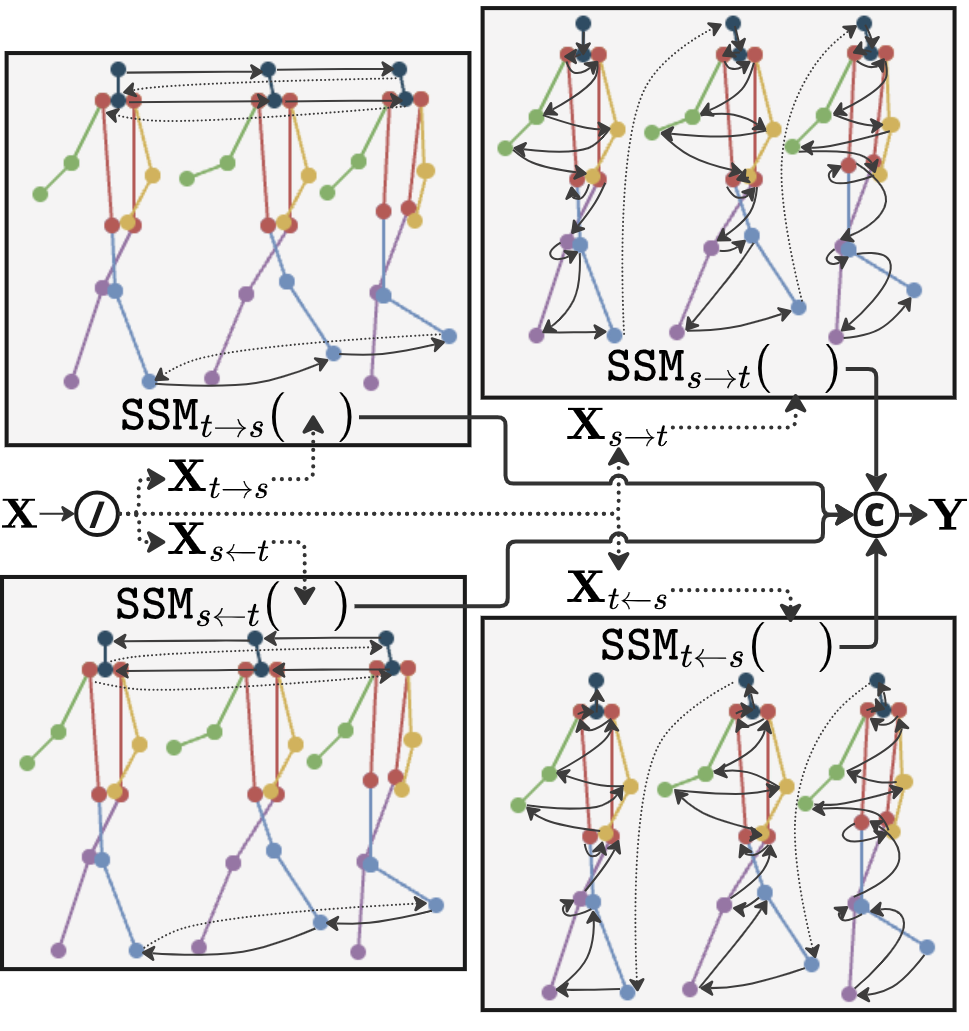}
\caption{Channel-Wise Spatio-Temporal SSM (C-2D-SSM).}
\label{fig:c2dssm}
\end{figure}
The Mamba architecture has been extended from 1D to 2D bidirectional modeling (\eg,~\cite{ma2024umamba,archit2024vimunet,xu2024mambasurvey}) showing promise in image-related tasks but exhibiting instability when scaled to large parameter spaces~\cite{patro2024simba}.
This is due to the Mamba block's extensive input and output projections, whose computational and parametric complexities scale linearly with the input channel dimensionality. 

To effectively mitigate these issues, we leverage the insight that different channel groups in skeletal data may represent distinct yet complementary aspects of movement.
As shown in~\figurename~\ref{fig:c2dssm}, we decompose the input $\mathbf{X}$ along the channel dimension to get equally sized tensors
\begin{equation}
\left[ \mathbf{X}_{t \rightarrow s}, \mathbf{X}_{s \rightarrow t}, \mathbf{X}_{t \leftarrow s}, \mathbf{X}_{s \leftarrow t} \right] = \mysplit(\mathbf{X})
\end{equation}
that are independently processed by direction-specific 2D-SSMs (\ie, spatial-temporal SSMs), then concatenated back to produce the final output
\begin{equation}
\mathbf{Y} = \myconcat
    \left(
        \begin{array}{cc}
\ssm_{t \rightarrow s}(\mathbf{X}_{t \rightarrow s}), \\
\ssm_{s \rightarrow t}(\mathbf{X}_{s \rightarrow t}), \\
\ssm_{t \leftarrow s}(\mathbf{X}_{t \leftarrow s}), \\
\ssm_{s \leftarrow t}(\mathbf{X}_{s \leftarrow t}) \\
    	\end{array}
    \right) = \mycssm(\mathbf{X})
\end{equation}
preserving the input tensor's dimensionality.
$t$ and $s$ denote temporal and spatial dimensions, respectively, and the arrows indicate the direction of spatial-temporal scanning. 

This novel parallel approach captures complementary movement features across channel groups, while significantly reducing the computational complexity by operating on $C/4$ channel inputs.
The direction-specific spatial-temporal scanning mechanism effectively captures diverse contextual information, enhancing the model's capacity to learn both local and global dependencies.

\subsubsection{Part-Grouped Modeling:}~
Joint locations change over time depending on the disease type.
Different deficits are caused by the involvement of multiple systems and structures (\eg, cerebellar, pyramidal, extrapyramidal).
For instance, hereditary paraplegia mostly involves lower body part joints~\cite{Serrao2016}, while Parkinson's and Cerbella Ataxia involve the whole body~\cite{10.1093/brain/122.7.1349}.
Motivated by these intuitions, we introduce both part-based and global SSMs to effectively capture fine-grained local motion details at individual body parts' level, while maintaining the global understanding of full-body motion. 


We decompose the body keypoints into multiple partitions corresponding to key body parts (arms, legs, torso) and their relevant combinations (arms-legs, arms-torso, torso-legs).
By focusing on these specific partitions, we enable our model to capture both localized movements and inter-segment temporal dynamics crucial for disease recognition.

For each partition $p \in \{1, \cdots, P\}$, we apply:
\begin{equation}
\mathbf{X}_p = \mycssm_p(\mathbf{b}_p + \mathbf{X}[:, \mathcal{I}_p])
\end{equation}
where $\mathcal{I}_p$ is the index set, $\mathbf{b}_p \in \mathbb{R}^{T \times |\mathcal{I}_p| \times C}$ is a learnable partition token, and $\mycssm_p$ is our novel C-2D-SSM for partition $p$.

The learnable partition token $\mathbf{b}_p$ enables each SSM to learn a specialized representation of the motion dynamics specific to each body part or combination, enhancing the model's ability to differentiate between fine-grained movements.
By processing these partitions separately, our model can efficiently capture part-specific temporal patterns, such as the rhythmic alternation of arm and leg movements during walking, or the stabilizing role of the torso in maintaining balance.

To ensure that full-body motion patterns are not missed when focusing solely on individual parts, we capture holistic motion characteristics as $\mathbf{X}_g = \mycssm(\mathbf{X})$.

\subsubsection{Attentive SSM:~}
The outputs from these specialized SSMs are then integrated through a learnable weighted sum:
\begin{equation}
\mathbf{X}_{\text{SSM}} = \sum_p \beta_p \mathbf{X}_p + \beta_g \mathbf{X}_g
\end{equation}
where $\beta_p$ and $\beta_g$ are learnable parameters.
This integration strategy allows our model to dynamically adjust the importance of part-specific and global motion information based on the input data.


To further refine our feature representations and enhance the model's adaptability, we incorporate a channel attention mechanism after the SSM processing.
We first compute
\begin{equation}
\mathbf{w} = \sigma(\linear(\gelu(\linear(\pool(\mathbf{X}_{\text{SSM}})))))
\end{equation}
where $\pool$ is the spatial-temporal average pooling operator and $\sigma$ is the sigmoid function.
Then, we obtain the output of our novel \pgml as
\begin{equation}
\pgml(\mathbf{X}) = (\mathbf{X}_{\text{SSM}} + \mathbf{X}) \odot \mathbf{w}
\end{equation}
where $\alpha$ is a learnable parameter and $\odot$ denotes the Hadamard product.
The residual connection ($+ \mathbf{X}$) ensures efficient gradient flow during training, while the channel attention ($\odot \mathbf{w}$) allows our model to adaptively recalibrate channel-wise feature responses~\cite{hu2018squeeze}, focusing on the most informative features for each input sequence.

%% file: sec/4_exp.tex
\section{Experiments}

\subsection{Datasets}

\subsubsection{Medical Diagnosis:}~Our objective is to provide a method for automated diagnosis of motion-related disorders.
To assess the capabilities of our method in such a context, we validate it on one newly collected dataset and a publicly available benchmark.
Both datasets present a challenging scenario for automated diagnosis from skeleton action recognition, requiring models to differentiate between multiple neurological conditions and healthy controls based on subtle motion characteristics.
\\
\textbf{Neurological Disorders (ND).}~
We collected a new dataset focused on automated diagnosis of neurological disorders.
Our dataset comprises 396 video sequences from 40 subjects across four distinct classes: primary degenerative cerebellar ataxia (11 patients, 112 sequences), hereditary spastic paraparesis (12 patients, 105 sequences), idiopathic Parkinson's disease (7 patients, 80 sequences), and healthy controls (10 subjects, 99 sequences). 
Data collection was conducted under carefully controlled conditions to ensure consistency and reliability. Each sequence captures a patient walking in a standardized environment, with multiple recordings per subject in both directions - towards and away from the camera.
All sequences were recorded using an HD camera operating at 30 fps.
The average sequence length is 140.64 frames, with a range from 69 to 465 frames, providing sufficient temporal context for analyzing motion characteristics associated with each condition.
\\
\textbf{KOA-PD-NM}~\cite{kour2024koanmpddataset} contains marker-based sequences of subjects with Knee Osteoarthritis (KOA, 50 patients), Parkinson's Disease (PD, 20 patients), and Normal (NM, 30 patients) motion patterns.
This dataset was captured in a controlled indoor environment (subjects walked on a green mat in the sagittal plane) using a 50fps HD camera.
The dataset includes varying severity levels (mild, moderate, severe) for KOA and PD subjects, graded by medical specialists.
We report the results computed considering and excluding the severity levels (\ie, yielding a 7-class and a 3-class dataset, respectively).  

\input{tables/sota_comparison_medical}

\subsubsection{Generic Actions:}~
While our approach is primarily designed for automated diagnosis of motion-related neurological disorders, we hypothesize its potential in generic action recognition tasks. To validate this hypothesis and provide a comprehensive evaluation, we conduct experiments on established public benchmarks for skeleton-based action recognition.
\\
\textbf{NTU RGB+D}~\cite{shahroudy2016ntu60} comprises 60 action classes, encompassing a wide range of daily activities.
It contains 56,880 video samples captured from 40 subjects across 155 camera viewpoints using Kinect v2, providing RGB, IR, depth, and 3D skeleton data. The dataset supports cross-subject (X-Sub60) and cross-view (X-View60) evaluation protocols.
11 classes involve two-person interactions, forming the NTU-Inter subset~\cite{duan2023skeletr,pang2022igformer,do2024skateformer}.
\\
\textbf{NTU RGB+D 120}~\cite{liu2020ntu120} extends the NTU RGB+D dataset to  120 action classes with 114,480 video samples from 106 subjects.
It introduces cross-subject (X-Sub120) and cross-setup (X-Set120) evaluation protocols.
The dataset includes 26 classes focused on human interaction, designated as NTU-Inter 120 \cite{duan2023skeletr,pang2022igformer,do2024skateformer}.
\\
\textbf{NW-UCLA}~\cite{wang2014nwucla} consists of 1,475 video samples spanning 10 action classes, performed by 10 subjects and captured from three distinct camera views. It provides RGB, IR, depth, and 3D skeleton data. Evaluation is conducted using a cross-view protocol~\cite{zhou2022hyperformer,do2024skateformer}, utilizing two views for training and one for testing.

\subsection{Implementation Details}
We trained our model on an NVIDIA L40S GPU using the PyTorch framework for 500 epochs on 128-sample batches, using the AdamW optimizer with weight decay equal to $5e-4$.
We adopted a linear learning rate warm-up from 1e-7 to 1e-3 over the first 25 epochs, followed by a cosine annealing scheduler.
We applied clipping for gradients having $\ell_2$-norm exceeding 1.
The label-smoothed cross-entropy loss with $\alpha= 0.1$ was used for optimization.
For the ND and KOA-PD-NM datasets, we used~\cite{sun2019hrnet} to obtain the skeleton joints.
For others, we used the provided keypoints.

\subsection{State-of-the-art Comparison}
We present a comprehensive performance comparison of our method against recent state-of-the-art skeleton-based action recognition methods.
Following~\cite{zhou2023, do2024skateformer,zhou2022hyperformer}, we considered three different modalities:
(i) only joints ($\modalityj$)
(ii) joints and bones ($\modalityjb$);
and (iii) joints and bones with motions ($\modalityjbm$).
We trained a model for each modality and ensembled their outputs.

\subsubsection{Medical Diagnosis:}~
Table~\ref{tab:sota_medical} provides a comparative evaluation of different models on three medical diagnosis datasets
Our proposed model achieves the highest accuracy across all datasets and ensemble metrics.
For the ND dataset, we reach 99.35\% on $\modalityj$ and $\modalityjb$, further improving to 99.64\%  on $\modalityjbm$, demonstrating robust performance in diagnosis of neurological disorders.

In the KOA-PD-NM dataset, we significantly improve other methods reading a 98.62\% accuracy using the $\modalityjbm$.
This indicates a strong capability in classifying grouped severity levels.
On the more challenging KOA-PD-NM-Severity dataset, our method better handles fine-grained severity distinctions outperforming all existing models.

Our novel state-space model architecture achieves substantial performance gains across datasets, highlighting its effectiveness for nuanced medical diagnosis tasks that require precise classification of disease severity.
This is likely due to the model's ability to dynamically capture long-range local and global joint interactions.

\subsubsection{Generic Actions:}
\input{tables/sota_comparison_generic}

\input{tables/sota_comparison_generic_inter}

Results in Table~\ref{tab:sota_generic} show that our noble method achieves leading performance across diverse generic action recognition benchmarks.
We consistently outperformed prior methods, demonstrating the effectiveness of our modeling approach.
On the NTU RGB+D X-Sub60 dataset, we recorded top scores of 91.8\%, 92.8\%, and 93.4\% across the three considered ensemble strategies, showing the capabilities of our approach in capturing spatial-temporal dependencies.
For the more complex NTU RGB+D 120 dataset, our method demonstrates excellent performance by improving over prior GCN and Transformer-based methods.
Similarly, on the NW-UCLA dataset, we obtain the highest accuracy at $97.6\%$ on the leaderboard.

Similar comments can be made for the results obtained comparing our method with state-of-the-art human interaction recognition methods on the NTU-inter and NTU-Inter 120 datasets.
Table~\ref{tab:sota_generic_inter} shows that we consistently achieve superior performance than existing methods, demonstrating the robustness and effectiveness of our approach in handling complex human interactions.

\subsubsection{Complexity Analysis:}
\input{tables/sota_complexity_analysis}
Table~\ref{tab:sota_complexity_analysis} provides a comparative computational complexity analysis with existing architectures.
Our approach achieves the highest accuracy, recording an average joint modality accuracy of 94.3\% and 88.1\%  on the NTU RGB+D and NTU RGB+D 120 datasets respectively --surpassing all existing methods.
While SkelMamba exhibits a slightly higher parameter count (6.84M) and FLOPs (9.7G) than some lighter GCN models, its inference time of 7.06 ms shows it is remarkably efficient. This trade-off positions our method as the most accurate and fastest model.

%% file: tables/sota_comparison_medical.tex
\begin{table*}[t]
\centering
\caption{Comparison with existing methods on the three considered medical diagnosis benchmark datasets: ND, KOA-PD-NM (3 classes with grouped severity levels), and KOA-PD-NM-Severity (7 classes with different diseases' severity levels).
Results are shown considering the number of input frames and ensemble strategies.}
\label{tab:sota_medical}
\footnotesize
\begin{tabularx}{\linewidth}{Y l Y Y Y Y Y Y Y Y Y}
\toprule
 & & \multicolumn{3}{c}{\textbf{ND}} & \multicolumn{3}{c}{\textbf{KOA-PD-NM}} & \multicolumn{3}{c}{\textbf{KOA-PD-NM-Severity}} \\
\cmidrule(lr){3-5} \cmidrule(lr){6-8} \cmidrule(lr){9-11}
 Type & Model & $\modalityj$ & $\modalityjb$ & $\modalityjbm$ & $\modalityj$ & $\modalityjb$ & $\modalityjbm$ & $\modalityj$ & $\modalityjb$ & $\modalityjbm$ \\
\midrule
CNN & PoseC3D~\cite{duan22posec3d} & 98.95 & 99.21 & 99.35 & 84.79 & 89.40 & 89.86 & 95.79 & 98.95 & 98.95 \\
\hline
\multirow{2}{*}{GCN} & 2S-AAGCN~\cite{shi2019agcn} & 94.74 & 97.89 & 98.95 & 91.22 & 92.39 & 93.18
 & 89.47 & 93.68 & 93.68 \\
 & STGCN++~\cite{yan2018stgcn} & 93.68 & 94.74 & 95.79 & 93.54 & 94.31 & 95.32 & 95.32 & 95.79 & 96.84 \\
\hline
Transformer & Hyperformer~\cite{zhou2022hyperformer} &  99.21 & 99.35 & 99.21 & 96.82 & 96.94 & 98.14 & 95.79 & 96.84 & 98.95 \\
\hline
\rowcolor{gray!25}
\cellcolor{white}
SSM & \textbf{SkelMamba (Ours)} & \textbf{99.35} & \textbf{99.35} & \textbf{99.64} & \textbf{97.45} & \textbf{98.23} & \textbf{98.62} & \textbf{96.84 }& \textbf{98.95} & \textbf{99.21} \\
\bottomrule
\end{tabularx}
\end{table*}

%% file: tables/sota_comparison_generic.tex
\begin{table*}[ht]
\centering
\caption{Comparison with existing methods on the three considered generic actions benchmark datasets: NTU RGB+D, NTU RGB+D 120, and NW-UCLA.
Results are shown considering the number of input frames and ensemble strategies.}
\label{tab:sota_generic}
\footnotesize
\begin{tabularx}{\linewidth}{Y l Y YYY YYY YYY YYY Y}
\toprule
\multirow{3}{*}{Types} & \multirow{3}{*}{Methods} & \multirow{3}{*}{Frames} & \multicolumn{6}{c}{\textbf{NTU RGB+D}} & \multicolumn{6}{c}{\textbf{NTU RGB+D 120}} & \textbf{NW} \\
 & & & \multicolumn{3}{c}{X-Sub60} & \multicolumn{3}{c}{X-View60} & \multicolumn{3}{c}{X-Sub120} & \multicolumn{3}{c}{X-Set120} & \textbf{UCLA}\\
  \cmidrule(lr){4-6} \cmidrule(lr){7-9}  \cmidrule(lr){10-12} \cmidrule(lr){13-15} \cmidrule(lr){16-16}   
                       &                          &                         & $\modalityj$   & $\modalityjb$   & $\modalityjbm$   & $\modalityj$   & $\modalityjb$   & $\modalityjbm$   & $\modalityj$   & $\modalityjb$   & $\modalityjbm$   & $\modalityj$   & $\modalityjb$   & $\modalityjbm$   &                           \\ 
                       \midrule
\multirow{1}{*}{RNN}   & AGC-LSTM~\cite{si2019agclstm}                 & 100                      & 87.5 & 89.2 & -    & 93.5 & 95.0 & -    & -    & -    & -    & -    & -    & -    & 93.3                      \\ 
\hline
\multirow{2}{*}{CNN}   & TA-CNN~\cite{xu2022tacnn}                   & 64                      & 88.8 & -    & 90.4 & 93.6 & -    & 94.8 & 82.4 & -    & 85.4 & 84.0 & -    & 86.8 & 96.1                      \\
                       & Ske2Grid~\cite{cai2023ske2grid}                 & 100                       & 88.3 & -    & -    & 95.7 & -    & -    & 82.7 & -    & -  & 85.1   & -    & -    & -                         \\ 
                       \hline
\multirow{10}{*}{GCN}  & SGN~\cite{zhang2020sgn}                      & 20                      & -    & 89.0 & -    & -    & 94.5 & -    & 79.2 & -    & -    & 81.5 & -    & -    & -                         \\
                       & CTR-GCN~\cite{chen2021ctrgcn}                  & 64                       & 89.9 & -    & 92.4 & -    & -    & 96.8 & 84.9 & 88.7 & 88.9 & - & 90.1 & 90.6    & 96.5                      \\
                       & ST-GCN++~\cite{yan2018stgcn}                 & 100                      & 89.3 & 91.4 & 92.1 & 95.6 & 96.7 & 97.0 & 83.2 & 87.0 & 87.5 & 85.6 & 87.5 & 89.8 & -                         \\
                       & InfoGCN~\cite{chi2022infogcn}                  & 64                       & -    & -    & 92.7 & -    & -    & 96.9 & 85.1 & 88.5 & 89.4 & 86.3 & 89.7 & 90.7 & 96.6                      \\
                       & FR-Head~\cite{zhou2023}                  & 64                      & 90.3 & 92.3 & 92.8 & 95.3 & 96.4 & 96.8 & 85.5 & - & 89.5 & 87.3 &  -  & 90.9   & 96.8  \\
                       & Koopman~\cite{wang2023koopman}                  & 64                      & 90.2 & - & 92.9 & 95.2 & -    & 96.8 & 85.7 & -    & 90.0 & 87.4 & -    & 91.3 & 97.0                      \\
                       & LST~\cite{xiang2023lst}                      & 64                      & 90.2 & - & 92.9 & 95.6 & -    & 97.0 & 85.5 & -    & 89.9 & 87.0 & -    & 91.1 & 97.2                      \\
                       & HD-GCN~\cite{lee2023hdgcn}                   & 64                      & 90.6 & 92.4 & 93.0 & 95.7 & 96.6 & 97.0 & 85.7 & 89.1 & 89.8 & 87.3 & 90.6 & 91.2 & 96.9                      \\
                       & STC-Net~\cite{lee2023stcnet}                  & 64                      & - & 92.5 & 93.0 & - & 96.7    & 97.1 & - & 89.3 & 89.9 & - & 90.7 & 91.3 & 97.2                      \\ 
                        & BlockGCN~\cite{zhou2024blockgcn}                  & 64                      & 90.9 & - & 93.1 & 95.4    & - & 97.0 & 86.9 & - & \textbf{90.3} & 88.2 & - & 91.5 & 96.9                      \\ 
                       \hline
\multirow{4}{*}{Transf.} & DSTA-Net~\cite{shi2020dstanet}            & 128                      & -    & -    & 91.5 & -    & -    & 96.4 & -    & -    & 86.6 & -    & -    & 89.0 & -                         \\
                       & STST~\cite{zhang2021stst}                     & 128                      & -    & -    & 91.9 & -    & -    & 96.8 & -    & -    & -    & -    & -    & -    & -                         \\
                       & FG-STFormer~\cite{gao2022fgstformer}              & 128                      & -    & -    & 92.6 & -    & -    & 96.7 & -    & -    & 89.0 & -    & -    & 90.6 & 97.0                      \\
                       & Hyperformer~\cite{zhou2022hyperformer}              & 64                      & 90.7 & -    & 92.9 & 95.1 & -    & 96.5 & 86.6 & -    & 88.0 & 88.0 & -    & 91.3 & 96.7                      \\
                       \hline
\multirow{2}{*}{SSM} & Simba~\cite{patro2024simba}            & 64                      & 89.0    & 90.5    & - & 94.4    & 95.2    & - & 79.7    & -    & - & 86.3    & -    & - & 96.3                         \\

\rowcolor{gray!25}
\cellcolor{white}
& \textbf{SkelMamba (Ours)}              & 64                      & \textbf{91.8} & \textbf{92.8} & \textbf{93.4} & \textbf{96.8} & \textbf{97.1} & \textbf{97.4} & \textbf{87.1} & \textbf{89.4} & 89.9 &\textbf{ 89.1} & \textbf{90.7} & \textbf{91.5} & \textbf{97.6} \\
\bottomrule
\end{tabularx}%

\end{table*}

%% file: tables/sota_comparison_generic_inter.tex
\begin{table*}[ht]
\centering
\caption{Comparison with existing human interaction recognition methods on the NTU-Inter
and NTU-Inter 120 datasets.}
\label{tab:sota_generic_inter}
\footnotesize
\begin{tabularx}{\linewidth}{Y l Y Y Y Y Y Y Y}
\toprule
\multirow{2}{*}{Types} & \multirow{2}{*}{Methods} & \multicolumn{2}{c}{NTU-Inter ($\modalityj$, \%)} & \multicolumn{2}{c}{NTU-Inter 120 ($\modalityj$, \%)} & \multirow{2}{*}{Params [M]} & \multirow{2}{*}{FLOPs [G]} & \multirow{2}{*}{Time [ms]} \\
\cline{3-4} \cline{5-6}
 & & X-Sub60 & X-View60 & X-Sub120 & X-Set120 & & & \\
\hline
\multirow{3}{*}{Transformer} & IGFormer~\cite{pang2022igformer} & 93.6 & 96.5 & 85.4 & 86.5 & - & - & - \\
 & SkeleTR~\cite{duan2023skeletr} & 94.9 & 97.7 & 87.8 & 88.3 & 3.82 & 7.30 & - \\
 & ISTA-Net~\cite{wen2023istanet} & - & - & 90.6 & 91.7 & 6.22 & 68.18 & 21.71 \\
 
\hline
 \rowcolor{gray!25}
\cellcolor{white}
  SSM                     & \textbf{SkelMamba (Ours)} & \textbf{96.5} & \textbf{98.9} & \textbf{92.0} & \textbf{92.8} & 6.84 & 9.7 & 7.06 \\

\bottomrule
\end{tabularx}%

\end{table*}

%% file: tables/sota_complexity_analysis.tex
\begin{table*}[t]
\centering
\caption{Computational analysis: comparison of parameters, FLOPs, inference time, and average top-1 accuracy for joint modality ($\uparrow$ higher is better, $\downarrow$ lower is better).}
\label{tab:sota_complexity_analysis}
\footnotesize
\begin{tabularx}{\linewidth}{Y l Y Y Y c c}
\toprule
Types & Methods & Params [M]$\downarrow$& FLOPs [G] $\downarrow$& Time [ms]$\downarrow$ & NTU RGB+D [\%]$\uparrow$& NTU RGB+D 120 [\%]$\uparrow$\\

\midrule
\multirow{5}{*}{GCN}   & InfoGCN~\cite{chi2022infogcn}              & 1.56                        & 3.34             & 12.97                      & -                               & 85.7                        \\ 
                       & FR-Head~\cite{zhou2023}             & 1.45              & 3.60                      & 18.49                      & 92.8                           & 86.4                        \\ 
                       & Koopman~\cite{wang2023koopman}             & 5.38                        & 8.76                      & 17.86                      & 92.7                             & 86.6                        \\ 
                       & LST~\cite{xiang2023lst}                & 2.10                        & 3.60                      & 18.85                      & 92.9                            & 86.3                        \\ 
                       & HD-GCN~\cite{lee2023hdgcn}              & 1.66                        & 3.44                      & 72.81                      & 93.2                             & 86.5                        \\ 
\hline
\multirow{2}{*}{Transformer} & DSTA-Net~\cite{shi2020dstanet}       & 3.45                        & 16.18                     & 13.80                      & -    & -                                                     \\ 

                       & Hyperformer~\cite{zhou2022hyperformer}         & 2.71                        & 9.64                      & 18.07                      & 92.9                            & 87.3                        \\ 
                       \hline
  \rowcolor{gray!25}
\cellcolor{white}
SSM                     &  \textbf{SkelMamba (Ours)} & 6.84 & 9.7 & 7.06 & \textbf{94.3} &  \textbf{88.1} \\
\bottomrule
\end{tabularx}%
\end{table*}

%% file: sec/5_ablation.tex
\subsection{Ablation Study}

\subsubsection{Part-Grouped Mamba Block Components:}~
\input{tables/ablation_components}
Table~\ref{tab:ablation_components} presents an ablation study on the novel part-grouped mamba block, investigating the contributions of its spatial, temporal, and spatio-temporal components.
The ablation of either the spatial or temporal component leads to a significant drop in accuracy, highlighting the importance of both modalities for the block's performance.
The PGM module, which models interactions of different parts of the body, consistently improves accuracy across all configurations.
This suggests that the PGM effectively captures the spatial relationships between body parts.

\subsubsection{Body partitioning and SSM scanning:}~
\input{tables/ablation_pgm}
Table~\ref{tab:ablation_pgm}~ presents the ablation study of the Part Grouped Mamba ($\pgml$) layer.
Transitioning from 1D SSM to C-2D-SSM notably improves performance, highlighting the benefit of spatial-temporal dependency modeling.
Incorporating attentive SSM enhances performance by capturing long-range dependencies, while channel attention refines feature representation, yielding additional gains.
Altogether, the fully integrated PGM layer achieves the highest accuracy, validating its design for robust part-based skeleton action recognition.

%% file: tables/ablation_components.tex
\begin{table*}[t]
\centering
\caption{Ablation analysis of the spatial, temporal, and spatio-temporal components of the novel part-grouped mamba block. Results show the top-1 accuracy for the $\modalityj$ modality ($\uparrow$ higher is better, $\downarrow$ lower is better).}
\label{tab:ablation_components}
\footnotesize
\begin{tabularx}{\linewidth}{Y Y Y l Y Y Y c}
\toprule

\multicolumn{3}{c}{Components} & \multicolumn{2}{c}{NTU RGB+D [\%] $\uparrow$} & \multirow{2}{*}{Params [M]$\downarrow$} & \multirow{2}{*}{FLOPs [G]$\downarrow$} & \multirow{2}{*}{Time [ms]$\downarrow$} \\ \cmidrule{1-3} \cmidrule{4-5}
       \gconv & \tconv & \pgml &  X-Sub60            & X-View60           &                             &                            &                                                        \\ 
       \midrule

    \tick & & & 89.9 & 94.6 & 4.3 &  8.2 & 2.65  \\
          & \tick  & & 87.2 & 93.5  & 4.1 & 8.1 & 1.80  \\
    & & \tick  &  90.9 & 95.1 & 6.6 & 9.5 & 6.90 \\
    \tick & \tick & & 90.6 & 94.8 & 4.3 & 8.2 & 2.99  \\
    \tick &  & \tick &  91.0 & 95.8 & 6.7 & 9.4 & 3.85  \\
     &  \tick & \tick &  91.1 &  95.4 & 6.6 & 9.4 & 5.76  \\
    \rowcolor{gray!25} \tick & \tick & \tick  & \textbf{91.8} & \textbf{96.8} & 6.8 & 9.7 & 7.06 \\

\bottomrule
\end{tabularx}%
\end{table*}

%% file: tables/ablation_pgm.tex
\begin{table}[t]
\centering
\caption{Contribution of the introduced components of the novel part grouped mamba (\pgml) layer. Results show the top-1 accuracy for the $\modalityj$ modality.}
\label{tab:ablation_pgm}
\footnotesize
\begin{tabularx}{\linewidth}{l Y Y}
\toprule

\multirow{2}{*}{\pgml} & \multicolumn{2}{c}{NTU RGB+D [\%]} \\ \cmidrule{2-3}
   &  X-Sub60            & X-View60  \\
   \midrule
1D SSM Baseline &  88.6 & 94.1 \\
+ 1D SSM to C-2D-SSM & 90.2 (+1.6)  & 95.3 (+1.2) \\
+ Partition Token ($\mathbf{b}_p$) & 90.5 (+0.3) & 95.6 (+0.3) \\
+ Attentive SSM ($\mathbf{X}_{\text{SSM}}$) & 91.6 (+1.1)  & 96.4 (+0.8) \\
+ Channel attention ($\mathbf{w}$) & \textbf{91.8} (+0.7) & \textbf{96.8} (+0.4) \\

\bottomrule
\end{tabularx}%
\end{table}

%% file: sec/6_conclusion.tex
\section{Conclusion}
We presented a novel skeleton-based action recognition framework that integrates anatomically-aware State Space Models (SSMs) for fine-grained analysis of spatio-temporal motion patterns. Our approach introduces a multi-stream architecture that partitions skeletal data into spatial, temporal, and spatio-temporal streams, enabling efficient and targeted analysis of complex human motions. Through anatomically-guided body part segmentation and a multi-directional scanning strategy, our method captures both local joint dynamics and global motion interactions crucial for applications requiring precision, such as automated medical diagnosis.

Extensive evaluations on standard benchmarks (NTU RGB+D, NTU RGB+D 120, NW-UCLA) demonstrate our model’s superiority by significantly improving over existing methods.
On a new medical dataset for gait analysis, our framework also exhibits strong potential for clinical diagnostics, accurately identifying subtle movement patterns indicative of neurological disorders.
However, our medical dataset remains limited in size, as expanding it requires complex data-gathering processes constrained by privacy concerns and regulatory restrictions. We are actively working to increase this dataset, but this endeavor remains resource-intensive.